\documentclass[letterpaper]{article} 
\usepackage{aaai24}  
\usepackage{times}  
\usepackage{helvet}  
\usepackage{courier}  
\usepackage[hyphens]{url}  
\usepackage{graphicx} 
\usepackage{natbib}  
\usepackage{caption} 
\usepackage{algorithm}
\usepackage{algorithmic}

\usepackage{newfloat}
\usepackage{listings}
\DeclareCaptionStyle{ruled}{labelfont=normalfont,labelsep=colon,strut=off} 
\floatstyle{ruled}
\newfloat{listing}{tb}{lst}{}
\floatname{listing}{Listing}
\usepackage{xspace}
\usepackage{bm}
\usepackage{amsmath}
\usepackage{booktabs}
\usepackage{subfigure}
\usepackage{multirow}
\usepackage{xcolor}

\definecolor{myblue}{rgb}{0.2, 0.2, 0.9}

\newcommand{\ours}{\textsc{SEType}\xspace}
\newcommand{\plm}{\textbf{PLM}}
\newcommand{\bmh}{\bm h}
\newcommand{\bmw}{\bm w}
\newcommand{\score}{\textrm{score}}
\newcommand{\etl}{\textrm{entail}}

\title{Seed-Guided Fine-Grained Entity Typing in Science and Engineering Domains}
\author{Yu Zhang$^{1}$\thanks{Equal Contribution.}, Yunyi Zhang$^{1*}$, Yanzhen Shen$^1$, \\ Yu Deng$^2$, Lucian Popa$^3$, Larisa Shwartz$^2$, ChengXiang Zhai$^1$, Jiawei Han$^1$}
\affiliations{
$^1$Department of Computer Science, University of Illinois at Urbana-Champaign, Urbana, IL, USA \\
$^2$IBM Thomas J. Watson Research Center, Yorktown Heights, NY, USA \\
$^3$IBM Almaden Research Center, San Jose, CA, USA \\
\{yuz9, yzhan238, yanzhen4, czhai, hanj\}@illinois.edu, \ \ \  \{dengy, lpopa, lshwart\}@us.ibm.com
}

\begin{document}

\maketitle

\begin{abstract}
Accurately typing entity mentions from text segments is a fundamental task for various natural language processing applications. Many previous approaches rely on massive human-annotated data to perform entity typing. Nevertheless, collecting such data in highly specialized science and engineering domains (e.g., software engineering and security) can be time-consuming and costly, without mentioning the domain gaps between training and inference data if the model needs to be applied to confidential datasets. In this paper, we study the task of \textit{seed-guided fine-grained entity typing} in science and engineering domains, which takes the name and a few seed entities for each entity type as the only supervision and aims to classify new entity mentions into both seen and unseen types (i.e., those without seed entities). To solve this problem, we propose \ours which first enriches the weak supervision by finding more entities for each seen type from an unlabeled corpus using the contextualized representations of pre-trained language models. It then matches the enriched entities to unlabeled text to get pseudo-labeled samples and trains a textual entailment model that can make inferences for both seen and unseen types. Extensive experiments on two datasets covering four domains demonstrate the effectiveness of \ours in comparison with various baselines. Code and data are available at: \url{https://github.com/yuzhimanhua/SEType}.
\end{abstract}

\section{Introduction}

Entity typing, i.e., automatically determining the types of entity mentions given their contexts, is a fundamental step for various text mining and natural language processing (NLP) tasks, such as entity linking \cite{wu2020scalable} and text classification \cite{hu2019heterogeneous}. 
Entity typing in science and engineering domains poses substantial new challenges, calling for dedicated research.
First, \textit{\underline{fine-grained entity typing} is critical for domain-specific applications}.
For example, in software and security domains (e.g., StackOverflow threads, GitHub README files, and vulnerability descriptions), entities need to be typed in fine-grained scale (e.g., devices, operating systems, versions, and functions) in order to drive downstream applications (e.g., technical question answering \cite{yu2020technical,yu2021technical} and knowledge graph construction \cite{rukmono2023enabling}).

Second, \textit{massive human annotation is too costly to be a solution.} 
Although entity typing has been extensively studied in NLP, most existing approaches rely on massive human-annotated training data, which are time-consuming and costly to obtain, especially in specialized technical domains. Moreover, practitioners in these domains often need to apply the model to their confidential datasets (e.g., internal software issue reports) which cannot be accessed by external annotators, incurring domain gaps between training and inference data. 
To alleviate annotation efforts, recent studies explore the setting of few-shot entity typing \cite{ding2022prompt,huang2022few,dai2023ultra}, where a few manually labeled samples are provided to train the model. However, unless the entity types are balanced (which is usually not the case -- the number of \textsc{Application} entities is 24 times more than the number of \textsc{Algorithm} entities in the StackOverflowNER dataset \cite{tabassum2020code}), one needs to sample and annotate a much larger number of entities to cover the minority types. 

\begin{figure*}[!t]
\centering
\includegraphics[width=\textwidth]{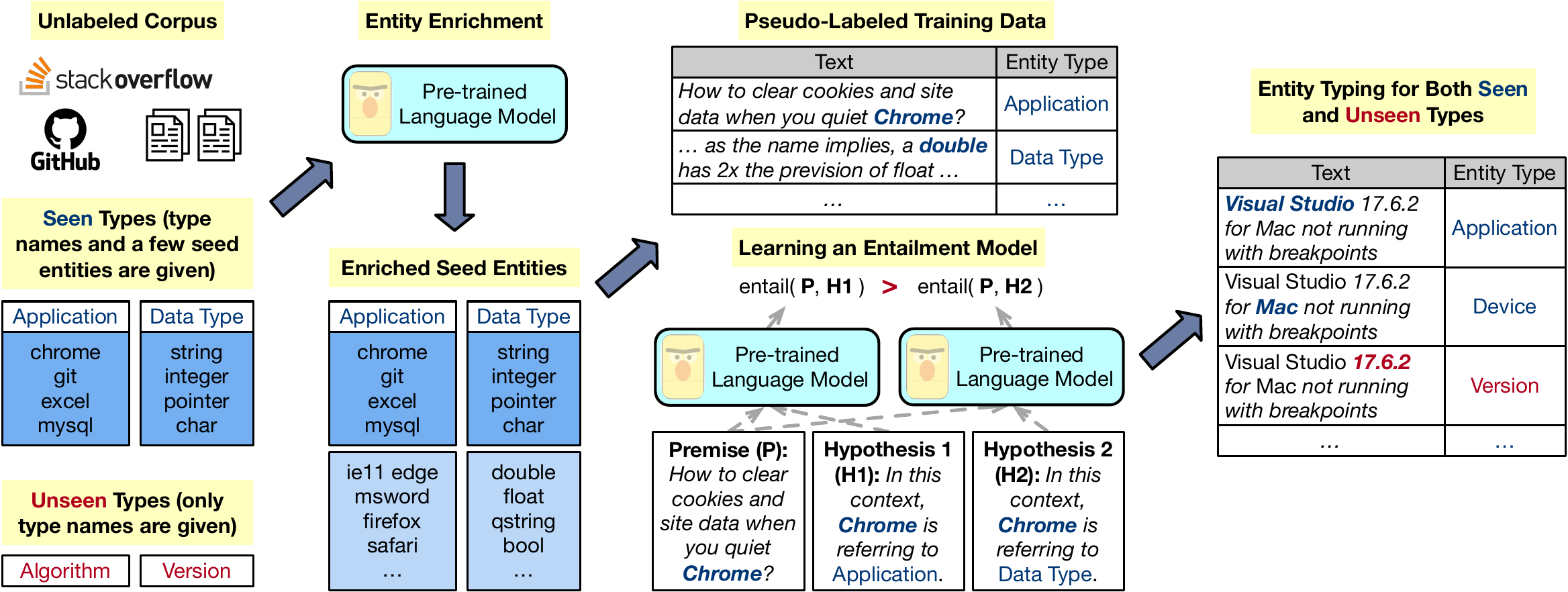}
\caption{Overview of the \ours framework.} 
\vspace{-1em}
\label{fig:framework}
\end{figure*}

Third, \textit{domain-agnostic zero-shot learning methods do not provide a good solution either.} 
Previous zero-shot entity typing models \cite{zhou2018zero,obeidat2019description,zhang2020mzet} rely on type names only and do not seek help from any annotated examples. This makes the model unaware of domain-specific knowledge, which may lead to suboptimal performance in highly specialized science and engineering domains. For example, given the sentence ``\textit{\underline{ListView} keep BooleanSparseArray of checked positions (you can get it with method getCheckedItemPositions()).}'', if we ask GPT-3.5 Turbo \cite{ouyang2022training} about the type of ``\textit{ListView}'' by providing it with type names only, GPT-3.5 Turbo will answer ``\textsc{User Interface Element}'' instead of the correct answer ``\textsc{Library Class}''.

To strike a balance between few-shot and zero-shot settings, in this paper, inspired by the weakly supervised setting in text classification \cite{meng2018weakly,mekala2020contextualized}, we confine the supervision signals to be type names and a few (e.g., 5) seed entities per type. In comparison with the labeled samples under the few-shot setting, the given entities under our setting are not associated with any context information. For example, the supervision only tells ``\textit{ListView}'' is a \textsc{Library Class} in general, but no specific sentence will be provided. In this case, users just need to name a few entities for each type as supervision, without scanning the text corpus. We call this task setting \textit{seed-guided entity typing}.

\vspace{1mm}
\noindent \textbf{Contributions.} In this paper, taking software engineering and security as two running domains, we study seed-guided fine-grained entity typing in science and technology. As shown in Figure \ref{fig:framework}, the task input includes a large unlabeled corpus and some \textit{seen} types. For each seen type, its type name and a small set of seed entities are given. The goal of our task is to train an entity typing model that can classify an entity mention into not only seen types but also \textit{unseen} types. By ``unseen'', we refer to the types without any seed entities given and never seen during model training. Instead, only their type names are provided during inference. For example, in Figure \ref{fig:framework}, \textsc{Version} is an unseen type, but the model should be able to type ``\textit{17.6.2}'' as a \textsc{Version} entity given its context ``\textit{Visual Studio \underline{17.6.2} for Mac not running with breakpoints}''.

To perform seed-guided fine-grained entity typing, we propose a framework named \ours, which consists of two phases: (i) entity enrichment and (ii) entailment model training. The first phase (i.e., entity enrichment) aims to extract more entities for each seen type from the unlabeled corpus according to their contextualized semantic similarities with the provided seed entities. This is to overcome supervision scarcity. To be specific, if we directly match seed entities to unlabeled text to get pseudo-labeled training samples, the semantic coverage of these obtained samples may still be narrow and cause model overfitting. After finding more entities belonging to each type, the matched training data will be more diverse. Taking such pseudo-labeled data, the model training phase then learns an entailment model by viewing the entity context as a premise and each entity type (filled into a template) as a hypothesis. Finally, the learned model can perform entity typing for both seen and unseen types by predicting to what extent the premise entails the hypothesis corresponding to each candidate type.

To summarize, this study makes the following contributions: (1) \textit{Task}: We propose to study seed-guided fine-grained entity typing. In comparison with the zero-shot setting, it leverages user-provided seed entities as domain knowledge, which is badly needed in highly specialized science and engineering domains. In comparison with the few-shot setting, it alleviates annotation efforts and mitigates domain gaps between training and inference data. (2) \textit{Framework}: We design a two-phase framework, \ours, that first conducts entity enrichment to overcome supervision scarcity and then learns an entailment model to perform entity typing for both seen and unseen types. (3) \textit{Experiments}: Extensive experiments on two public datasets \cite{tabassum2020code,bridges2013automatic} covering four domains (i.e., StackOverflow, GitHub, National Vulnerability Database, and Metasploit) demonstrate the effectiveness of \ours given 10 to 15 fine-grained types related to code, software, and security. Although we focus on software and security domain examples, our entity typing framework can be applied to other specialized domains including science \cite{wang2021chemner} and engineering \cite{o2021ms}.

\section{Problem Definition}
Assume there are $m$ entity types $\mathcal{T}=\{t_1,t_2,...,t_m\}$ (e.g., ``\textsc{Data Structure}'', ``\textsc{Device}'', ``\textsc{Programming Language}''). For each type $t_i$, a small set of (e.g., 5) seed entities $\mathcal{E}_i=\{e_{i,1},...,e_{i,n}\}$ are given as supervision (e.g., for $t_i=$ ``\textsc{Programming Language}'', $\mathcal{E}_i=\{\textit{c++, java, python, ...}\}$). The seed-guided entity typing task aims to train a classifier $f$. Given a text segment $d$ and an entity mention $e$ appears in $d$, the classifier maps $e$ to its type $f(e|d)$ based on its context. In this paper, we consider two different task settings.

\vspace{1mm}

\noindent \textbf{Closed-Set Entity Typing.} During inference, the possible type of an entity $e$ always belongs to $\mathcal{T}$. In other words, when making predictions, the model $f$ only needs to consider the entity types it has seen during training.

\vspace{1mm}

\noindent \textbf{Open-Set Entity Typing.} Following \citet{yuan2018otyper}, we consider a more challenging setting where some target entity types $\mathcal{T}_u=\{t_{m+1},t_{m+2},...,t_{m+k}\}$ are never seen during training. For each unseen type $t_i \in \mathcal{T}_u$, no seed entity is given as supervision. During inference (i.e., after the model $f$ is trained), the name of each unseen type (e.g., ``\textsc{Version}'') is given to describe it, and the possible type of an entity $e$ can be either seen or unseen.

Formally, our task is defined as follows.

\newtheorem{definition}{Definition}
\begin{definition}[Problem Definition]
Given $m$ entity types $\mathcal{T}=\{t_1,t_2,...,t_m\}$, each of which has its name $t_i$ and a small set of seed entities $\mathcal{E}_i=\{e_{i,1},...,e_{i,n}\}$ as input, \textbf{seed-guided entity typing} aims to train a classifier $f$ that maps an entity $e$ mentioned in a text segment $d$ to its type $f(e|d)$. Under the \textbf{closed-set} setting, $f(e|d) \in \mathcal{T}$; under the \textbf{open-set} setting, $f(e|d) \in \mathcal{T} \cup \mathcal{T}_u$, where $\mathcal{T}_u$ is a set of new types with no seed entities given and never seen during training.
\end{definition}

\section{The \ours Framework}
Since only a few seed entities are provided for each seen type, if we directly match them to an unlabeled corpus to get pseudo-labeled training data, the matched samples may still be scarce and cause model overfitting. Therefore, we propose a two-phase framework, \ours (shown in Figure \ref{fig:framework}), which first enriches each type with more entities and then trains an entailment-based entity typing model with enriched pseudo-labeled samples.

\subsection{Entity Enrichment}
In the first phase, given the sets of seed entities $\mathcal{E}_i=\{e_{i,1},...,e_{i,n}\}$ $(1\leq i\leq m)$, our goal is to find more entities $\mathcal{E}_i^+ =\{e_{i,n+1},...,e_{i,n+l}\}$ that also belong to type $t_i$ from a large unlabeled corpus $\mathcal{D}$. This subtask bears similarities with the entity set expansion task \cite{rong2016egoset,shen2017setexpan,yu2019corpus,zhang2020empower}. The difference is that here we need to expand the entity sets of multiple types simultaneously. To be specific, since the expanded entities will be used to match unlabeled corpus to derive pseudo-labeled training data, we would like them to be unambiguous (i.e., always belonging to the same type given different contexts) so that the obtained pseudo labels are likely more accurate. Therefore, we keep mutual exclusivity across different types during expansion (i.e., $(\mathcal{E}_i \cup \mathcal{E}_i^+) \cap (\mathcal{E}_j \cup \mathcal{E}_j^+) = \emptyset, \ \forall i \neq j$).

Given a large corpus $\mathcal{D}$ in a science or engineering domain, we first extract a candidate entity pool $\mathcal{P}$ from $\mathcal{D}$ that will be considered during entity enrichment. Following \citet{zhang2022entity}, we adopt AutoPhrase \cite{shang2018automated} to implement this step.
Then, we leverage a pre-trained language model $\plm$ (e.g., BERTOverflow \cite{tabassum2020code}, which is a BERT model pre-trained on software-related text corpora) to get a representation vector $\bmh_e$ for each seed entity $e \in \bigcup_{i=1}^m \mathcal{E}_i$ or candidate entity $e \in \mathcal{P}$. To achieve this, we find all sentences from $\mathcal{D}$ that contain the entity $e$. For each such sentence $d$, following \citet{zhang2022unsupervised}, we consider two ways to obtain the contextualized embedding of $e$: (1) We directly feed $d$ into $\plm$. Note that the entity $e$ may be segmented into multiple tokens by $\plm$. After encoding, each token in $d$ will have an embedding vector, and the embedding of $e$ is the average embedding of tokens in $e$. (2) We replace $e$ with a \texttt{[MASK]} token in $d$ and feed the masked sentence into $\plm$. After encoding, we view the embedding of \texttt{[MASK]} as the embedding of $e$. One can observe that the former approach mentioned above focuses more on the \textit{content} of $e$, while the latter emphasizes the \textit{context} of $e$. We concatenate these two embeddings as the sentence-level representation of $e$ in $d$. Finally, the corpus-level representation of $e$ is the average of all its sentence-level representations in $\mathcal{D}$. Formally,
\begin{equation}
    \bmh_e = \frac{1}{{\rm sf}(e, \mathcal{D})} \sum_{d} \Big[\bmh_{e|d}^{\rm content}\ ||\ \bmh_{e|d}^{\rm context}\Big].
\label{eqn:embedding}
\end{equation}
Here, ${\rm sf}(e, \mathcal{D})$ is the number of sentences containing $e$ in $\mathcal{D}$ (i.e., ``sentence frequency''); $\bmh_{e|d}^{\rm content}$ and $\bmh_{e|d}^{\rm context}$ are the sentence-level representations of $e$ in $d$ calculated in the two aforementioned ways.

After obtaining $\bmh_e$, we conduct an iterative entity enrichment process for each type. In each iteration, whether a candidate entity $e$ should be added to $\mathcal{E}^+_i$ is according to the average cosine similarity between $e$ and all entities already belonging to type $t_i$. Formally,
\begin{equation}
    \score(e, t_i) = \frac{1}{|\mathcal{E}_i \cup \mathcal{E}_i^+|}\sum_{e'\in\mathcal{E}_i \cup \mathcal{E}_i^+} \cos(\bmh_e, \bmh_{e'}).
\label{eqn:score}
\end{equation}

At the very beginning of the iterative process, we have $\mathcal{E}^+_i=\emptyset$ $(1\leq i\leq m)$. After each iteration, for each type $t_i$, we sort all candidate entities according to $\score(e, t_i)$, and the top-ranked entities will be added to $\mathcal{E}^+_i$. To ensure mutual exclusivity, each entity can only be added to its most similar type. For example, if $\score(e, \textsc{Library}) = 0.8$ and $\score(e, \textsc{Application}) = 0.7$, then $e$ will not be added to the \textsc{Application} type, even if it is top-ranked according to $\score(e, \textsc{Application})$. Equivalently, we define the following score.
\begin{equation}
   \score'(e, t_i)= 
   \begin{cases}
       \score(e, t_i), & \textrm{if } t_i = \arg\max_t \score(e, t) \\
       0, & \textrm{otherwise}.
   \end{cases}
\label{eqn:score_new}
\end{equation}
For each type $t_i$, the ranking criterion is
\begin{equation}
    \max_e \score'(e, t_i), \textrm{ where } e \in \mathcal{P}\backslash \big(\bigcup_{j=1}^m (\mathcal{E}_j \cup \mathcal{E}^+_j)\big).
\end{equation}

\subsection{Entailment Model Training}
After $I$ iterations of entity enrichment, we have a set of entities $\mathcal{E}_i \cup \mathcal{E}^+_i$ belonging to each type $t_i$. We use these entities to match an unlabeled corpus. If there is a sentence $d$ in the corpus containing any entity $e \in \mathcal{E}_i \cup \mathcal{E}^+_i$, we create a pseudo-labeled training sample denoted by the triplet $(e, d, t_i)$, which means that an entity $e$ is mentioned in its context $d$ and should be labeled as type $t_i$. Observing that some sentences in software-related text corpora (e.g., StackOverflow QA threads) are succinct and do not contain sufficient context information, we propose to include the $\pm c$ context sentences of $d$ also as the input to the entity typing model. For the sake of brevity, we use $d$ to denote the text segment consisting of both the sentence mentioning $e$ and its context sentences.

Following \citet{li2022ultra}, we propose to train a natural language inference (NLI) model \cite{yin2019benchmarking} for entity typing. To be specific, given a pseudo-labeled training sample $(e, d, t_i)$, we treat $d$ as a \textit{premise}, fill $e$ and $t_i$ into a template ``\textit{In this context, \underline{$e$} is referring to \underline{$t_i$}.}'' to construct a \textit{hypothesis}, and train the model to recognize that the premise entails the hypothesis. Similar to existing NLI models such as RoBERTa-large-MNLI \cite{liu2019roberta} and BART-large-MNLI \cite{lewis2020bart}, our model adopts a Cross-Encoder architecture, which concatenates the premise and the hypothesis into one input sequence and feeds it into $\plm$.
\begin{equation}
    \bmh_{\texttt{[CLS]}} = \plm(\texttt{[CLS]}\ d\ \texttt{[SEP]}\ \eta(e, t_i)\ \texttt{[SEP]}).
\label{eqn:cross1}
\end{equation}
Here, $\eta(e, t_i)$ denotes the hypothesis; $\bmh_{\texttt{[CLS]}}$ is the output representation vector of \texttt{[CLS]}.
We then stack a linear layer upon $\plm$ to predict to what extent the hypothesis is correct given the premise.
\begin{equation}
    \etl(d, \eta(e, t_i)) = \bmw^\top \bmh_{\texttt{[CLS]}},
\label{eqn:cross2}
\end{equation}
where $\bmw$ is a trainable vector.

According to the pseudo label, $e$ belongs to type $t_i$ rather than any other type $t_j \in \mathcal{T}\backslash\{t_i\}$. Therefore, the premise $d$ should entail the hypothesis $\eta(e, t_i)$ and should not entail $\eta(e, t_j)$. To encourage this, we utilize a contrastive loss \cite{smith2005contrastive} during training:
\begin{equation}
\begin{split}
    \mathcal{J} & = -\sum_{t_i \in \mathcal{T}} \sum_{(e, d, t_i)} \sum_{t_j \in \mathcal{T}\backslash\{t_i\}} \log \Big( \\
    & \frac{\exp(\etl(d, \eta(e, t_i)))}{\exp(\etl(d, \eta(e, t_i))) + \exp(\etl(d, \eta(e, t_j)))} \Big).
\end{split}
\end{equation}
The model parameters, including $\plm$ and $\bmw$, are trained on all pseudo-labeled samples derived by $\mathcal{E}_i \cup \mathcal{E}^+_i (1\leq i \leq m)$ to minimize $\mathcal{J}$.

\subsection{Inference}
After $\plm$ and $\bmw$ are trained, given a testing sample $(e, d)$, we are able to predict the type of $e$. In brief, we enumerate all possible hypotheses to see which one is the most likely to be entailed by $d$. Under the closed-set setting, the hypothesis space is $\mathcal{H}=\{\eta(e, t_j)|t_j \in \mathcal{T}\}$; under the open-set setting, the hypothesis space becomes $\mathcal{H}=\{\eta(e, t_j)|t_j \in \mathcal{T}\cup\mathcal{T}_u\}$. Formally, for each $\eta(e, t_j) \in \mathcal{H}$, we calculate $\etl(d, \eta(e, t_j))$ according to Eqs. (\ref{eqn:cross1}) and (\ref{eqn:cross2}). Then, we pick the type with the highest entailment score as the predicted type of $e$.
\begin{equation}
    f(e|d) = \arg\max_{t_j: \eta(e, t_j) \in \mathcal{H}} \etl(d, \eta(e, t_j)).
\end{equation}

\section{Experiments}
\subsection{Datasets}
We use two publicly available datasets from software engineering and security domains -- StackOverflowNER \cite{tabassum2020code} and Cybersecurity \cite{bridges2013automatic}.

\vspace{1mm}
\noindent \textbf{StackOverflowNER} \cite{tabassum2020code}\footnote{\url{https://github.com/jeniyat/StackOverflowNER}} contain text from two sources -- \textbf{StackOverflow} question-answer threads and \textbf{GitHub} issue reports. We select 10 types as seen types defined in our seed-guided setting, including \textsc{Application}, \textsc{Data Structure}, \textsc{Data Type}, \textsc{Device}, \textsc{Library}, \textsc{Library Class}, \textsc{Operating System}, \textsc{Programming Language}, \textsc{User Interface Element}, and \textsc{Website}, and we pick 5 types as unseen types, including \textsc{Algorithm}, \textsc{File Type}, \textsc{HTML XML Tag}, \textsc{Value}, and \textsc{Version}. In the original dataset, StackOverflow question-answer threads are split into training, validation, and testing sets, while GitHub issue reports form a testing set only. We take the two testing sets as our testing sets, which are named  \textbf{StackOverflow} and \textbf{GitHub}, respectively. We take the training and validation corpora of StackOverflow, remove their annotations, and treat them as unlabeled corpora to create pseudo-labeled training and validation sets, respectively.

\vspace{1mm}
\noindent \textbf{Cybersecurity} \cite{bridges2013automatic}\footnote{\url{https://github.com/stucco/auto-labeled-corpus}} contains text related to the Common Vulnerability Enumeration (CVE) from the National Vulnerability Database (\textbf{NVD}) and the \textbf{Metasploit} Framework. We select 5 types as seen types -- \textsc{Application}, \textsc{Edition}, \textsc{Operating System}, \textsc{Relevant Term}, and \textsc{Vendor}; 5 other types are treated as unseen types -- \textsc{File}, \textsc{Function}, \textsc{Method}, \textsc{Parameter}, and \textsc{Version}. For the larger \textbf{NVD} corpus, we take 20\% as the annotated testing data, and the remaining 80\% are treated as unlabeled text to create pseudo-labeled training and validation data. For the smaller \textbf{Metasploit} corpus, we take all annotated samples for testing.

\vspace{1mm}
For each seen type, 4-7 seed entities are given. Statistics of the datasets are summarized in Table \ref{tab:data}.

The large unlabeled corpus $\mathcal{D}$ for entity enrichment is sampled from the Stack Exchange data dump\footnote{\url{https://archive.org/download/stackexchange/stackoverflow.com-Posts.7z}}. $\mathcal{D}$ consists of 1.26 million questions and answers.

\begin{table}[!t]
\small
\centering
\begin{tabular}{c|cc|cc}
\toprule
\multirow{3}{*}{\textbf{Dataset}} & \multicolumn{2}{c|}{\textbf{StackOverflowNER}} & \multicolumn{2}{c}{\textbf{Cybersecurity}} \\
\cmidrule{2-5}
 & \begin{tabular}[c]{@{}c@{}}\textbf{StackOv-} \\ \textbf{erflow} \end{tabular}        & \textbf{GitHub}        & \textbf{NVD}           & \begin{tabular}[c]{@{}c@{}}\textbf{Meta-} \\ \textbf{sploit} \end{tabular}        \\
\midrule
\#Seen types                                                             & 10                   & 10            & 5             & 5                 \\
\midrule
\#Unseen types                                                           & 5                    & 4$^\dagger$             & 5             & 5                 \\
\midrule
\begin{tabular}[c]{@{}c@{}}Average \#seeds \\ per seen type \end{tabular} & 5.4                & 5.4         & 5.2        & 5.2             \\
\midrule
\begin{tabular}[c]{@{}c@{}}\#Testing samples\\ (Closed-Set)\end{tabular} & 2,084                & 3,224         & 20,798        & 2,087             \\
\midrule
\begin{tabular}[c]{@{}c@{}}\#Testing samples\\ (Open-Set)\end{tabular}   & 2,610                & 3,762         & 27,646        & 2,612            \\
\bottomrule
\end{tabular}
\caption{Dataset statistics. $\dagger$: The GitHub dataset does not have any entity annotated as \textsc{Value}.}
\vspace{-0.5em}
\label{tab:data}
\end{table}

\begin{table*}[!t]
\small
\centering
\begin{tabular}{c|cccc|cccc}
\toprule
 & \multicolumn{4}{c|}{\textbf{StackOverflow}}                            & \multicolumn{4}{c}{\textbf{GitHub}}                                   \\
 & \multicolumn{2}{c}{Closed-Set}   & \multicolumn{2}{c|}{Open-Set}     & \multicolumn{2}{c}{Closed-Set}   & \multicolumn{2}{c}{Open-Set}     \\
 & Micro           & Macro           & Micro           & Macro           & Micro           & Macro           & Micro           & Macro           \\
\midrule
RoBERTa-large-MNLI \cite{liu2019roberta}    & 34.74$^{**}$          & 31.30$^{**}$          & 27.85$^{**}$          & 24.57$^{**}$          & 34.62$^{**}$          & 29.87$^{**}$          & 31.55$^{**}$          & 25.65$^{**}$          \\
BART-large-MNLI \cite{lewis2020bart}       & 30.95$^{**}$          & 24.31$^{**}$          & 27.20$^{**}$          & 21.08$^{**}$          & 23.39$^{**}$          & 22.09$^{**}$          & 24.46$^{**}$          & 21.75$^{**}$          \\
BERTOverflow \cite{tabassum2020code}         & 27.59$^{**}$          & 25.34$^{**}$          & --               & --               & 20.47$^{**}$          & 17.73$^{**}$          & --               & --               \\
SetExpan+Entailment \cite{shen2017setexpan} & 48.99$^{**}$          & 48.06$^{**}$          & 40.96$^{**}$          & 35.00$^{**}$          & 39.14$^{**}$          & 47.16$^{**}$          & 35.73$^{**}$          & 36.72$^{**}$          \\
CGExpan+Entailment \cite{zhang2020empower} & 61.37$^{**}$          & 59.97$^{*}$          & 56.02$^{**}$          & 49.43$^{**}$          & 45.97$^{**}$          & 52.13          & 47.45$^{*}$          & 46.85$^{*}$          \\
GPT-3.5 Turbo \cite{ouyang2022training} & 50.79$^{**}$          & 48.07$^{**}$          & 51.86$^{**}$          & 49.23$^{**}$          & 45.49$^{**}$          & 46.37$^{**}$          & 49.85          & 48.42 \\
\midrule
\ours & \textbf{66.15} & \textbf{64.16} & \textbf{60.05} & \textbf{52.83} & \textbf{52.30} & \textbf{55.20} & \textbf{52.45} & \textbf{49.83} \\
\midrule
Fully Supervised      & 82.77          & 82.85          & 73.52          & 65.62          & 74.63          & 77.51          & 71.66          & 65.24         \\
\bottomrule
\toprule
 & \multicolumn{4}{c|}{\textbf{NVD}}                            & \multicolumn{4}{c}{\textbf{Metasploit}}                                   \\
 & \multicolumn{2}{c}{Closed-Set}   & \multicolumn{2}{c|}{Open-Set}     & \multicolumn{2}{c}{Closed-Set}   & \multicolumn{2}{c}{Open-Set}     \\
 & Micro           & Macro           & Micro           & Macro           & Micro           & Macro           & Micro           & Macro           \\
\midrule
RoBERTa-large-MNLI \cite{liu2019roberta}   & 64.00$^{**}$          & 29.34$^{**}$          & 52.55$^{**}$          & 29.53$^{**}$          & 56.06$^{**}$          & 26.92$^{**}$          & 51.34$^{**}$          & 24.93$^{**}$          \\
BART-large-MNLI \cite{lewis2020bart}      & 65.43$^{**}$          & 20.66$^{**}$          & 51.26$^{**}$          & 18.12$^{**}$          & 59.61$^{**}$          & 23.55$^{**}$          & 54.42$^{**}$          & 21.67$^{**}$          \\
BERTOverflow \cite{tabassum2020code}         & 26.07$^{**}$          & 15.46$^{**}$          & --               & --               & 31.77$^{**}$          & 20.56$^{**}$          & --               & --               \\
SetExpan+Entailment \cite{shen2017setexpan} & 66.86$^{**}$ & \textbf{57.07} & 51.01$^{**}$ & 29.27$^{**}$ & 71.70 & \textbf{60.15} & 57.40$^{**}$ & 31.52 \\
CGExpan+Entailment \cite{zhang2020empower} & 58.63$^{**}$ & 53.10 & 49.64$^{**}$ & 32.35 & 67.86$^{**}$ & 59.48 & 57.09$^{**}$ & 34.61 \\
GPT-3.5 Turbo \cite{ouyang2022training} & 54.71$^{**}$ & 46.41$^{**}$ & 50.45$^{**}$ & \textbf{45.25} & 60.04$^{**}$ & 50.43$^{**}$ & 50.19$^{**}$ & \textbf{42.78} \\
\midrule
\ours & \textbf{75.93} & 55.34 & \textbf{61.69} & 36.62 & \textbf{74.77} & 57.26 & \textbf{62.50} & 34.72 \\
\midrule
Fully Supervised & 97.67 & 90.50 & 74.17 & 39.47 & 98.28 & 82.09 & 78.87 & 40.58 \\
\bottomrule
\end{tabular}
\caption{Performance of compared methods on StackOverflow and GitHub from the StackOverflowNER dataset \cite{tabassum2020code} as well as NVD and Metasploit from the Cybersecurity dataset \cite{bridges2013automatic}. Bold: the highest score. *: \ours is significantly better than this method with p-value $< 0.05$. **: \ours is significantly better than this method with p-value $< 0.01$.}
\vspace{-0.5em}
\label{tab:main}
\end{table*}

\subsection{Compared Methods}
We compare \ours with the following baselines.

\vspace{1mm}
\noindent \textbf{RoBERTa-large-MNLI} \cite{liu2019roberta} is a natural language inference model which is obtained by fine-tuning RoBERTa-large on the MNLI dataset \cite{williams2018broad}. \citet{li2022ultra} propose to utilize it as a zero-shot entity typing model by viewing the mention text as a premise and the candidate entity type as a hypothesis.

\vspace{1mm}
\noindent \textbf{BART-large-MNLI} \cite{lewis2020bart} is obtained by fine-tuning BART-large on MNLI. It can be used as a zero-shot entity typing model in the same way as RoBERTa-large-MNLI.

\vspace{1mm}
\noindent \textbf{BERTOverflow} \cite{tabassum2020code} is a base-size BERT model pre-trained on StackOverflow data. We adopt it for few-shot entity typing. To be specific, for each entity mention $e$ appearing in text $d$, we feed $d$ into BERTOverflow and get the contextualized embedding of $e$ (i.e., $\bmh_{e|d}^{\rm content}$ in Eq. (\ref{eqn:embedding})). Then, for each seen type $t_i$, we obtain its embedding by feeding its seed entities $\mathcal{E}_i$ into BERTOverflow (without any context) and taking the average of their embeddings. Finally, we select the type that is the most similar to $e$ in the BERTOverflow embedding space (measured by the cosine similarity) as the prediction. Note that this method can only be used for closed-set entity typing because unseen types do not have seed entities.

\vspace{1mm}
\noindent \textbf{SetExpan+Entailment} first uses SetExpan \cite{shen2017setexpan} for entity enrichment and then uses enriched entities to create pseudo-labeled training data to train an entailment model. The entailment model is initialized with BERTOverflow. It can be viewed as an ablation version of \ours by replacing our entity enrichment step with SetExpan.

\vspace{1mm}
\noindent \textbf{CGExpan+Entailment} is similar to SetExpan+Entail- ment, but it replaces our entity enrichment step with CGExpan \cite{zhang2020empower}.

\vspace{1mm}
\noindent \textbf{GPT-3.5 Turbo} \cite{ouyang2022training} is a large language model pre-trained on massive corpora with instructions and human feedback. We use it for zero-shot entity typing by inputting the entity $e$ and the context $d$ and asking the model to select an entity type from the candidate type space.

\subsection{Implementation and Hyperparameters}
\ours uses BERTOverflow as the \plm. During entity enrichment, the number of enriched entities $|\mathcal{E}_i^+|$ is 50 and 100 on StackOverflowNER and Cybersecurity, respectively. In practice, this hyperparameter can be set based on users' knowledge about the rough number of concepts belonging to the types. During model training, the window size of context sentences $c=1$; the maximum premise length is 462 tokens; the maximum hypothesis length is 50 tokens; the training batch size is 4; we use the AdamW optimizer \cite{loshchilov2019decoupled}, warm up the learning rate for the first 100 steps and then linearly decay it, where the learning rate is 5e-5; the weight decay is 0.01, and $\epsilon =$ 1e-8. The model is trained on one NVIDIA RTX A6000 GPU.

\subsection{Evaluation Metrics}
We use Micro-F1 and Macro-F1 scores as evaluation metrics for both closed-set and open-set settings.

\subsection{Performance Comparison}
Table \ref{tab:main} shows the performance of compared methods on StackOverflowNER and Cybersecurity, respectively. We run \ours five times with the average performance reported. To show statistical significance, we conduct a two-tailed Z-test to compare \ours with each baseline, and the significance level is also shown in Table \ref{tab:main}. We also present the performance of a fully supervised entity typing model, where the ground-truth training and validation sets from StackOverflow and NVD are used to train an entailment model. Note that the term ``fully supervised'' here corresponds to the closed-set setting rather than the open-set one. In other words, the training and validation sets contain annotations for seen types only, while for unseen types, the model still only has their names during inference.

From Table \ref{tab:main}, we can observe that: (1) \ours outperforms all baselines significantly in most cases. On StackOverflow and GitHub, \ours is consistently the best. On NVD and Metasploit, \ours achieves the highest Micro-F1 scores and the second highest Macro-F1 scores. (2) If we do not perform any entity enrichment and directly utilize BERTOverflow for few-shot entity typing, the performance is even lower than the zero-shot RoBERTa-large-MNLI and BART-large-MNLI models in most cases. This is possibly because RoBERTa-large-MNLI and BART-large-MNLI are fine-tuned on general-domain NLI data and enjoy a larger PLM backbone. By contrast, SetExpan+Entailment, CGExpan+Entailment, and \ours add an entity enrichment step before using BERTOverflow, and they can outperform RoBERTa-large-MNLI and BART-large-MNLI in many cases. This observation underscores the importance of our proposed entity enrichment phase. (3) \ours beats SetExpan+Entailment and CGExpan+Entailment in most cases. Since the entailment model training modules of all these three models are identical, this finding implies that our proposed entity enrichment method is better than SetExpan and CGExpan. The possible reason is that our method is specifically designed for expanding multiple entity sets simultaneously and explicitly models mutual exclusivity across all types.

\subsection{Hyperparameter Analyses}

\begin{figure}[!t]
\centering
\subfigure[StackOverflow]{
\includegraphics[width=0.225\textwidth]{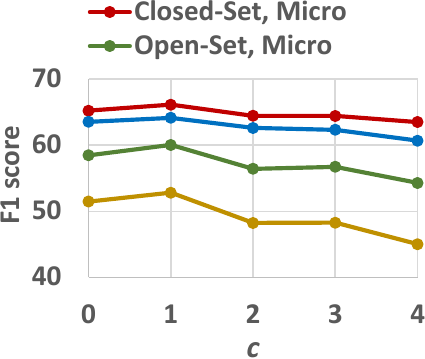}}
\subfigure[GitHub]{
\includegraphics[width=0.225\textwidth]{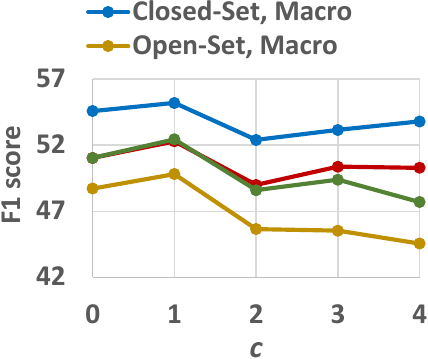}}
\vspace{-0.5em}
\caption{Performance of \ours with different window sizes of context sentences (the sentence containing the entity mention and its $\pm c$ sentences are fed into the PLM) on StackOverflow and GitHub. Considering $\pm 1$ sentences is always better than focusing on the mention sentence alone.} 
\label{fig:window}
\end{figure}

\noindent \textbf{Effect of the Window Size of Context Sentences.} As mentioned in our model design, we include the $\pm c$ context sentences of the mention sentence as model input to complement the information in the mention sentence. In \ours, we set $c=1$. Figure \ref{fig:window} depicts the performance of \ours with different values of $c$. We can find that: (1) The F1 scores of \ours with $c=1$ are consistently better than those with $c=0$. This trend validates our motivation for including context sentences. (2) If we further increase the value of $c$, the performance starts to fluctuate or even drop. This observation is intuitive because when we stretch too far away, the sentence may be irrelevant to the mentioned entity and bring more noises than hints.

\begin{figure}[!t]
\centering
\subfigure[StackOverflow]{
\includegraphics[width=0.225\textwidth]{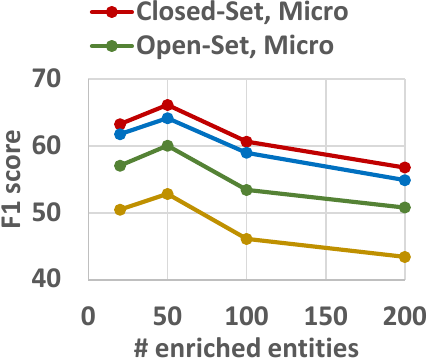}}
\subfigure[NVD]{
\includegraphics[width=0.225\textwidth]{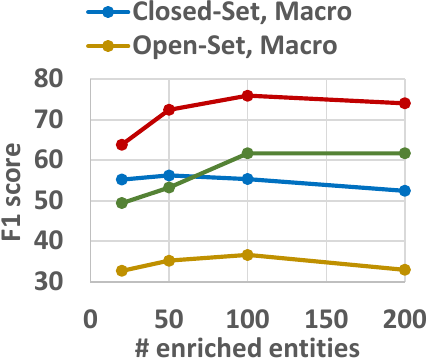}}
\vspace{-0.5em}
\caption{Performance of \ours with different numbers of enriched entities per type (i.e., $|\mathcal{E}_i^+|$) on StackOverflow and NVD.} 
\vspace{-0.5em}
\label{fig:expan}
\end{figure}

\vspace{1mm}
\noindent \textbf{Effect of the Number of Enriched Entities.} On StackOverflowNER and Cybersecurity, we extract 50 and 100 entities, respectively, for each seen type to form the enriched entity set $\mathcal{E}_i^+$. We now examine the effect of $|\mathcal{E}_i^+|$ on model performance, which is plotted in Figure \ref{fig:expan}. We can see that: (1) When we increase $|\mathcal{E}_i^+|$ from 20 to 50, \ours consistently performs better. This again implies the positive contribution of finding more entities to tackle supervision scarcity. (2) If we further increase $|\mathcal{E}_i^+|$ from 100 to 200, the performance often drops. This is because lower-ranked entities may not belong to the corresponding type, inducing errors and noises in pseudo-labeled training data. Moreover, extracting more entities will produce more training samples and make training time longer. Overall, we believe setting $|\mathcal{E}_i^+|$ to be 50 or 100 strikes a good balance.

\begin{table*}[!t]
\small
\centering
\begin{tabular}{c|cc|cc|cc|cc}
\toprule
\multirow{2}{*}{\begin{tabular}[c]{@{}c@{}} \textbf{Premise:} Visual Studio 17.6.2 for \underline{\textbf{Mac}} not\\ running with breakpoints\end{tabular}}  & \multicolumn{2}{c|}{\textbf{StackOverflow}} & \multicolumn{2}{c|}{\textbf{GitHub}} & \multicolumn{2}{c|}{\textbf{NVD}} & \multicolumn{2}{c}{\textbf{Metasploit}} \\
 & Closed               & Open                & Closed           & Open             & Closed          & Open           & Closed             & Open               \\
\midrule
\begin{tabular}[c]{@{}c@{}} \textbf{Contextual Template:} In this context, \underline{\textbf{Mac}} \\ is referring to \underline{\textsc{Device}}. \end{tabular}                    & \textbf{66.15}               & \textbf{60.05}              & \textbf{52.30}           & \textbf{52.45}           & \textbf{75.93}          & 61.69         & \textbf{74.77}             & 62.50             \\
\midrule
\textbf{Taxonomic Template:} \underline{\textbf{Mac}} is a \underline{\textsc{Device}}. & 65.29               & 58.85              & 51.59           & 50.63           & 75.42          & \textbf{63.64}         & 73.27             & \textbf{63.44} \\
\midrule
\begin{tabular}[c]{@{}c@{}} \textbf{Substitution Template:} Visual Studio 17.6.2 \\ for \underline{\textsc{Device}} not running with breakpoints \end{tabular} & 57.50$^{**}$               & 41.67$^{**}$              & 48.36           & 36.80$^{**}$           & 63.57$^{**}$          & 47.14$^{**}$         & 47.24$^{**}$             & 37.23$^{**}$ \\
\bottomrule
\end{tabular}
\caption{Micro-F1 scores of \ours with different hypothesis templates used in \citet{li2022ultra}. Bold, *, and **: the same meaning as in Table \ref{tab:main}.}
\vspace{-0.5em}
\label{tab:template}
\end{table*}

\vspace{1mm}
\noindent \textbf{Effect of the Hypothesis Template.} In \ours, we use the template ``\textit{In this context, \underline{entity} is referring to \underline{type}.}'' to create hypotheses. This template is introduced in \citet{li2022ultra}, where two other templates are also proposed. We show the three templates in Table \ref{tab:template} together with an example premise. Following the terminologies in \citet{li2022ultra}, we name the three templates ``Contextual'', ``Taxonomic'', and ``Substitution'', respectively. The Micro-F1 scores of \ours with different templates are also shown in Table \ref{tab:template}. We observe that the Substitution template is significantly inferior to the Contextual template; the Taxonomic template performs slightly worse than the Contextual template in general, but the gap is not significant.



\section{Related Work}
\subsection{Zero-shot and Few-shot Entity Typing}
Zero-shot and few-shot entity typing aim to classify a given entity mention to a set of types with limited or no training data. For example, ZOE \cite{zhou2018zero} and DZET \cite{obeidat2019description} propose to utilize pre-trained word embeddings (i.e., ELMo \cite{peters2018deep} and GLoVe \cite{pennington2014glove}, respectively) and Wikipedia’s entry descriptions to map entity mentions and types into a shared latent space; 
MZET \cite{zhang2020mzet} captures semantic meanings and hierarchical information to transfer knowledge from seen to unseen types; PLET \cite{ding2022prompt} uses prompt-learning with self-supervision to utilize a PLM’s prior knowledge; \citet{huang2022few} exploit a PLM's generative power to synthesize new instances for each entity type by feeding a prompt into a PLM and predicting the \texttt{[MASK]} tokens; \citet{cui2022prototypical} and \citet{yuan2022generative} continue to exploit the power of prompt tuning, while introducing a prototypical verbalizer or adding on prompt and curriculum instructions; \citet{ouyang2023ontology} exploit ontology structures of entity types and present an ontology enrichment framework for zero-shot entity typing. In addition, the power of transfer learning with PLMs has also been explored. For example, \citet{dai2023ultra} propose to train a general ultra-fine entity typing model and fine-tune it on fine entity typing data. 
However, in previous studies, context-aware annotated samples (i.e., labeled mentions with the sentence/document they appear in) are given under the few-shot setting. In comparison, in our setting, only a few seed entities are given, without their context, which are weaker supervision signals. In addition to the difference in the level of supervision, the previous studies aim to perform entity typing in the general domain, whereas ours works for specialized technical domains.

\subsection{Text Mining in Technical Domains}
Information extraction and text mining in technical domains such as software engineering and security have been explored to benefit domain-specific applications such as technical question answering \cite{yu2020technical,yu2021technical}, knowledge graph construction \cite{rukmono2023enabling}, and code recommendation \cite{jin2023code}. For named entity recognition (NER) in technical domains, \citet{ye2016software} show that conditional random fields can outperform traditional rule-based models; \citet{tabassum2020code} further leverage the power of PLMs and achieve better performance on StackOverflow and GitHub; \citet{lopez2021mining} find that the layout information from raw PDFs can help capture document structure for better NER performance. 
However, unlike \ours, these methods are all limited by their fully supervised settings requiring massive annotated data. 
In cases where annotations are insufficient or unavailable, \citet{bridges2013automatic} manage to automatically generate labeled data through matching entries in a security database; \citet{yang2021few} show that high-quality NER labels can be produced by PLMs based on security vulnerability reports. 
In comparison with the settings of these studies, \ours uses only a few seed words as supervision, which alleviates the burden of modeling knowledge bases or annotating training samples. 
For text classification in technical domains, \citet{zhang2019higitclass,zhang2020minimally,zhang2021hierarchical} devise weakly supervised approaches that use metadata and/or label hierarchy to classify GitHub repositories. While classification and entity expansion mainly require document or corpus-level understanding of text, the more challenging task of seed-guided entity typing, as described by \ours, relies on more fine-grained understanding of the context.

\section{Conclusions and Future Work}
In this paper, we study seed-guided fine-grained entity typing in science and engineering domains, which aims to perform entity typing given only type names and a small set of seed entities. 
We present \ours, a two-phase entity typing framework that first conducts entity enrichment and then employs the enriched entities to obtain pseudo-labeled data for subsequent entailment model training.
\ours is able to classify new entity mentions into both seen and unseen types. 
With extensive experiments on two public datasets encompassing multiple domains, we demonstrate the significant advantage of \ours over zero-shot and seed-guided baselines given 10 to 15 fine-grained types related to code, software, and security. Ablation studies and hyperparameter analyses further validate some key design choices in \ours.
Interesting future studies include: (1) leveraging large language models to synthesize pseudo-labeled training samples with the help of prompts and (2) exploiting domain-specific knowledge bases to create distant supervision to help fine-grained entity typing.

\section*{Acknowledgments}
We thank anonymous reviewers for their valuable and insightful feedback. This work was supported by the IBM-Illinois Discovery Accelerator Institute, National Science Foundation IIS-19-56151, and US DARPA INCAS Program No. HR001121C0165.

\bibliography{aaai24}
\end{document}